%% file: main.tex
\documentclass{article}

\usepackage[utf8]{inputenc}
\usepackage{lipsum}
\usepackage{graphicx}
\usepackage[backend=biber,style=numeric,sorting=none]{biblatex}
\usepackage{url}
\usepackage{hyperref}
\usepackage{amsmath}
\usepackage{subfig}
\usepackage{float}
\usepackage{tabularx}

\hypersetup{
    colorlinks=true,
    linktoc=all,
    linkcolor=black,
}

\begin{document}
    
    \include{frontpage}

    \tableofcontents

    \include{intro}

    \include{overview}

    \include{init}

    \include{detection}

    \include{pose}

    \include{illegal}

    \include{stats}

    \include{gui}

    \include{dataset}

    \include{results}
    
    \include{conclusions}
    
    \include{code}
    
	\printbibliography

\end{document}

%% file: frontpage.tex
\begin{table}[h!]
    \centering 
    \begin{tabular}{c c c}
        & 
        \begin{minipage}{.3\textwidth}
            \includegraphics[width=1.1\textwidth]{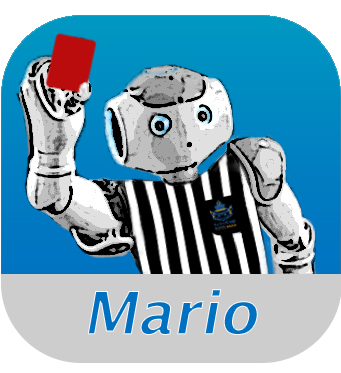}
            \vspace{2.5 mm}
        \end{minipage}
        &
        \\
        
        \begin{minipage}{.2\textwidth}
            \includegraphics[width=1.1\textwidth]{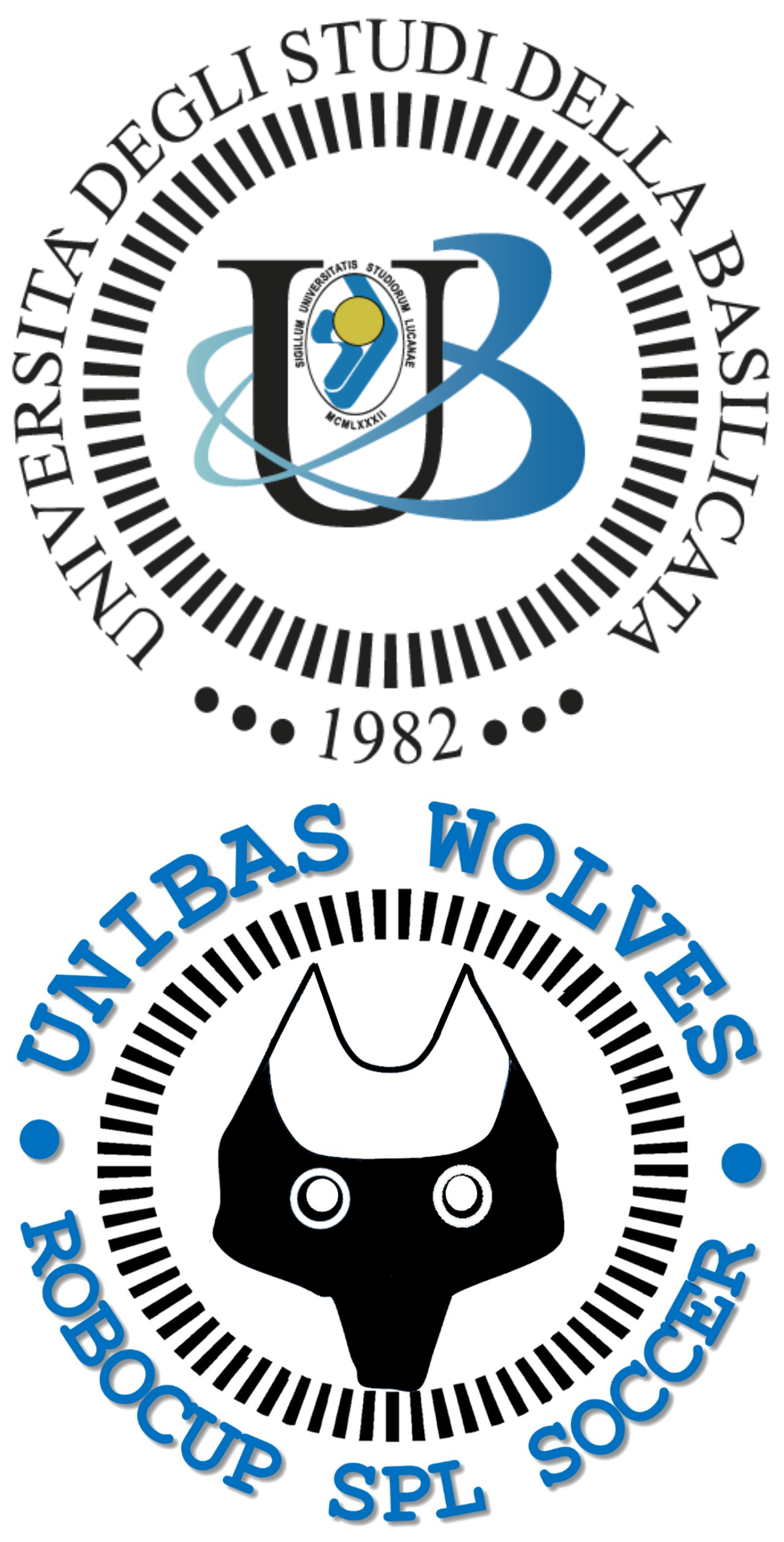}
        \end{minipage}
        &
        \begin{minipage}{7.2cm}
            \begin{center}
                \Large \textbf{MARIO: Modular and Extensible Architecture for Computing Visual Statistics in RoboCup SPL} \\
                \vspace{5 mm}
                \normalsize Domenico D. Bloisi, Andrea Pennisi, \\
                Cristian Zampino, Flavio Biancospino, \\
                Francesco Laus, Gianluca Di Stefano, \\
                Michele Brienza, and Rocchina Romano
            \end{center}
        \end{minipage}
        & 
        \begin{minipage}{.2\textwidth}
            \includegraphics[width=1.1\textwidth]{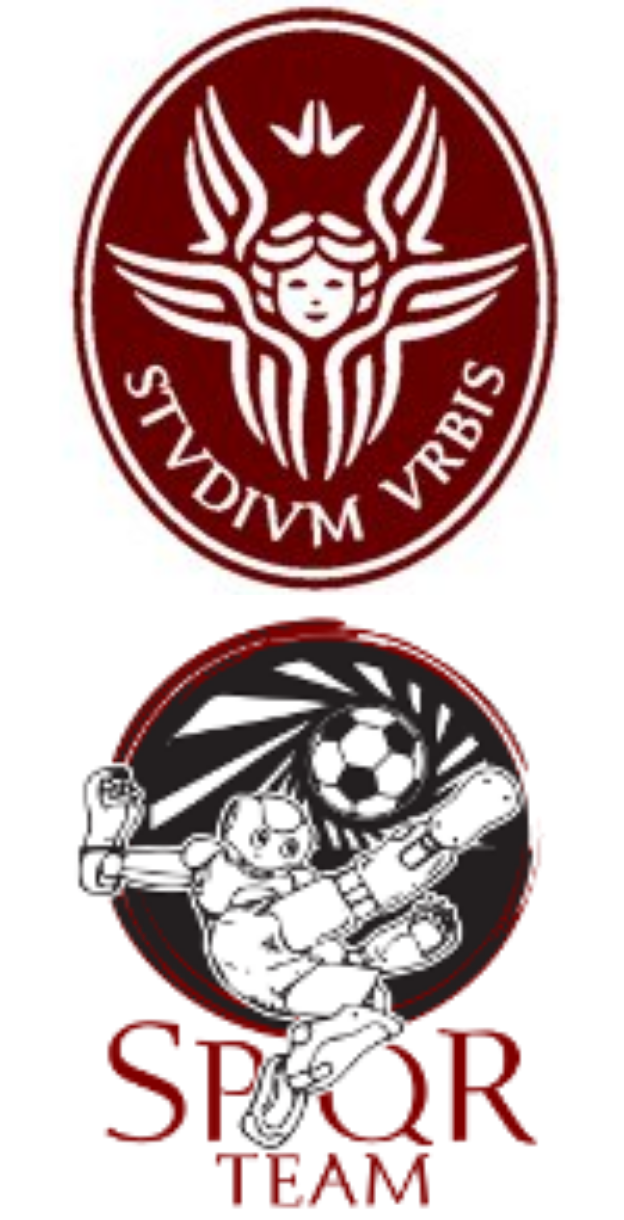}
        \end{minipage}
    \end{tabular}
\end{table}

\begin{abstract}
This technical report describes a modular and extensible architecture for computing visual statistics in RoboCup SPL (MARIO), presented during the SPL Open Research Challenge at RoboCup 2022, held in Bangkok (Thailand). MARIO is an open-source, ready-to-use software application whose final goal is to contribute to the growth of the RoboCup SPL community. MARIO comes with a GUI that integrates multiple machine learning and computer vision based functions, including automatic camera calibration, background subtraction, homography computation, player + ball tracking and localization, NAO robot pose estimation and fall detection. MARIO has been ranked no. 1 in the Open Research Challenge.

\end{abstract}

%% file: intro.tex
\section{Introduction} \label{intro}

    \subsection{RoboCup}

        RoboCup \cite{robocup_website} is an annual International robotics initiative founded in 1996 by a group of university professors. The main goal of the competition is to promote robotics and AI research by offering an appealing and formidable challenge. To this end, the competition set a long-term goal to create a fully autonomous humanoid robot team capable of competing and winning a soccer game, in compliance with the official rules of FIFA, by 2050.
        

    \subsection{Open Research Challenge}
    
        Since its foundation, one of the objectives of RoboCup  was to push the boundaries of research by offering high-level challenges. For the year 2022, the Open Research Challenge \cite{RoboCupTechnicalCommittee2022} proposed by the RoboCup Standard Platform League (often referred to as RoboCup SPL) focused on generating statistics of the matches from external videos captured by using a GoPro-like camera. Mellmann \textit{et al.}\cite{Mellmann2018} tried to extract statistics using only the \textit{GameController}/\textit{TeamCom} data without success. They realized that the data was not sufficient to extract consistent statistics from a match. For such a reason, the Open Research Challenge focuses on the use of a camera for extracting match statistics. 
        
        The challenge has 2 main goals:
        \begin{enumerate}
            \item to calculate extrinsic camera parameters (camera matrix) from the camera feed and to locate/track all moving objects (i.e.: ball and robots) on the field (short-term goals);
            \item to create game statistics (e.g.: time under control, successful, unsuccessful shots on goal, passes, etc.) based on the located objects and positions (long-term goals).
        \end{enumerate}
        
        

    \subsection{MARIO}
    
        MARIO is an end-to-end architecture for computing visual statistics in RoboCup SPL. One of the features of MARIO is the \textit{modular architecture}, which allows the user to customize the system for extracting statistics.
        
        UNIBAS WOLVES (the SPL team of the University of Basilicata) in collaboration with SPQR (the team of the Sapienza University of Rome)   presented MARIO at the Open Research Challenge at RoboCup 2022 held in Bangkok (Thailand). MARIO ranked as no. 1 ex-aequo with the B-Human team (see Table \ref{table:orc_2022_ranking}).
        
        \begin{table}[H]
            \centering
            \begin{tabular}{|c|c|c|} 
                \hline
                \textbf{Rank} & \textbf{Team} & \textbf{Score} \\ \hline
                1   & B-Human \& SPQR & 25   pt \\ 
                3   & NomadZ          & 21.9 pt \\ 
                4   & Nao Devils      & 18.1 pt \\
                5   & RoboEireann     & 13.5 pt \\
                6   & R-ZWEI KICKERS  & 5    pt \\
                \hline
            \end{tabular}
            \caption{Open-Research Challange 2022 ranking.}
            \label{table:orc_2022_ranking}
        \end{table}
        
        The final rank has been decided using a vote among all the SPL teams, which had the possibility to choose the five best teams through a board counting mechanism. The vote has been evaluated according to the following criteria:
        
        \begin{itemize}
            \item \textit{Achievement of long/short term goals};
            \item \textit{Execution time and hardware requirements};
            \item \textit{Metrics (accuracy/precision/recall)};
            \item \textit{Technical strength};
            \item \textit{Novelty}.
        \end{itemize}
        
    \subsection{Report Structure}
        
        The following sections provide the description of the system as well as experimental results. The remainder of the report is organized as follows:
        
        \begin{itemize}
            \item \hyperref[overview]{\texttt{MARIO System Overview}} which describes the architecture of the system;
            \item \hyperref[init]{\texttt{Initialization}} containing an overview of the preliminary operations performed by the software, such as camera calibration, background subtraction, and homography computation;
            \item \hyperref[detection]{\texttt{Detection, Tracking and Localization}} presents the detection and tracking systems;
            \item \hyperref[pose]{\texttt{Pose Estimation and Fall Detection}} and \hyperref[illegal]{\texttt{Illegal Defender}} show how the pose estimation systems works, and how the player and the illegal defender fouls are detected;
            \item \hyperref[stats]{\texttt{Data Association and Statistics}} describing how the visual data is merged with the data from the \textit{GameController}/\textit{TeamCom};
            \item \hyperref[gui]{\texttt{GUI}} and \hyperref[results]{\texttt{Experimental Results}}: give an overview of the graphical interface and the obtained results;
            \item \hyperref[conclusions]{\texttt{Conclusions}}: draws the conclusions and and future work.
            
        \end{itemize}

%% file: overview.tex
\section{MARIO System Overview} \label{overview}

    \subsection{Architecture}
    
\begin{figure}[h]
            \centering
            \includegraphics[scale=0.1]{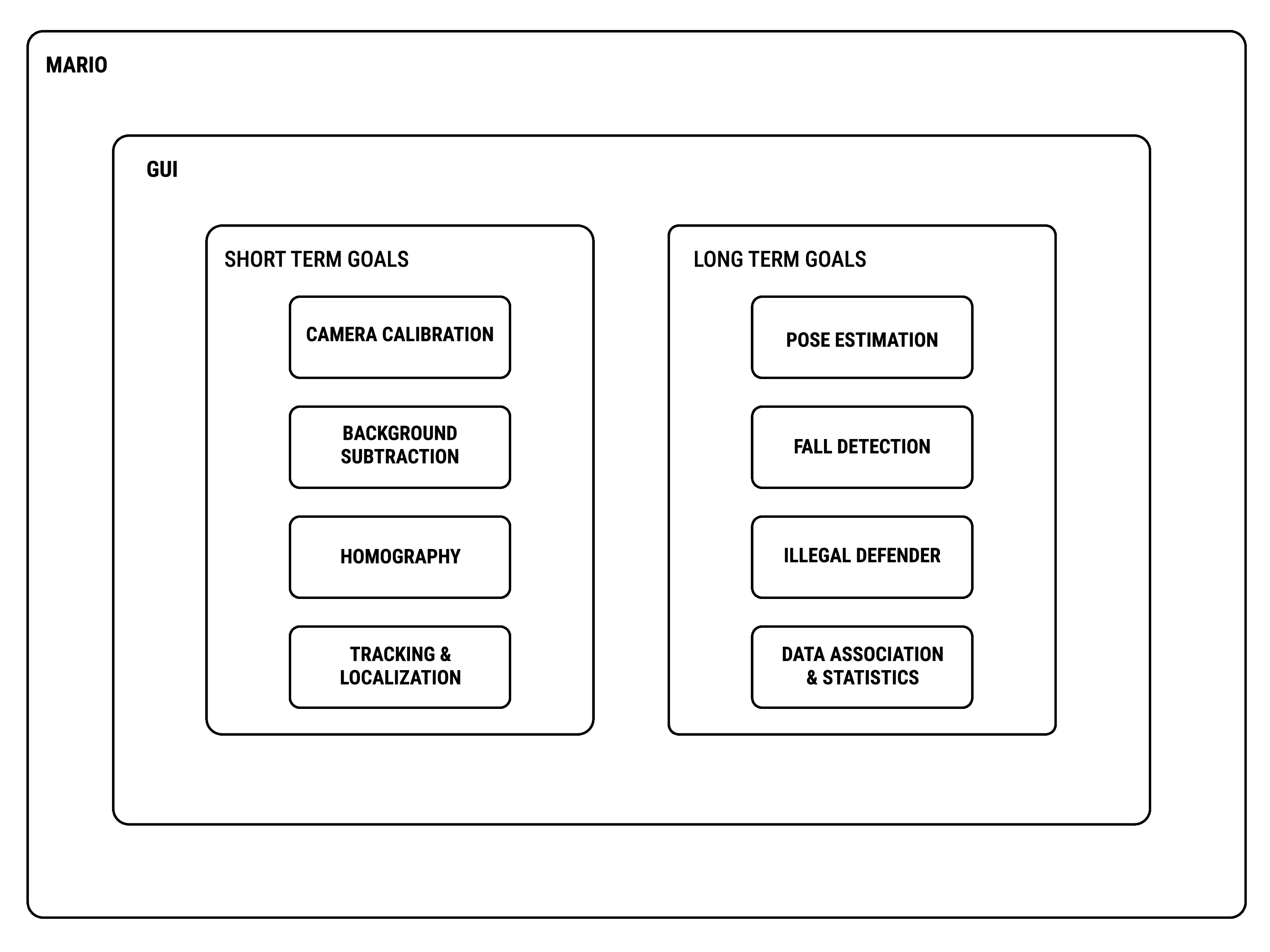}
            \caption{Architecture diagram of MARIO.}
            \label{fig:MARIO_architecture_diagram}
        \end{figure}

         MARIO has a modular architecture where each module can perform a specific task independently. Fig. \ref{fig:MARIO_architecture_diagram} shows the architecture diagram of MARIO. 
        
        Each module is built according to the goal pursued, whether belonging to one of the short-term or long-term goals. The short-term modules can be grouped as follows:
        
        \begin{itemize}
            \item \textit{Camera Calibration.} A preliminary camera calibration is performed in order to remove the camera lens distortion;
            \item \textit{Background Subtraction.} A background subtraction technique is applied for extracting the moving objects;
            \item \textit{Homography.} Homography is performed to compute a plan view of the field;
            \item \textit{Tracking and Localization.} YOLOv5 and StrongSORT models are used to track and localize the players and the ball.
        \end{itemize}
        
        While the long-term modules are:
        \begin{itemize}
            \item \textit{Pose Estimation.} A Convolutional Neural Network (CNN) model is used to compute the robot pose. A custom data-set is created specifically for such a task;
            \item \textit{Fall Detection.} Given the skeleton information, a Spatial-Temporal Graph Convolutional Networks (ST-GCN) model is used to detect the fall detection;
            \item \textit{Illegal Defender.} The tracking results are used to check if no more than 3 players from the same team are in the same penalty area;
            \item \textit{Data Association and Statistics.} Game data containing player and ball information is extracted and used to compute the statistics of the game.
        \end{itemize}

%% file: init.tex
\section{Initialization} \label{init}

    \subsection{Camera Calibration} \label{sec:camera_calibration}
    
        A \textit{camera calibration} procedure is performed in order to remove the distortion of the camera lens. In order to undistort the image, the \textit{intrinsic parameters} are required. Such parameters include information such as the \textit{focal length} $(f_x, f_y)$ and the \textit{optical center} $(o_x, o_y)$, which are used to calculate the \textit{camera matrix}. Such a matrix is a unique $3 \times 3$ matrix used to remove the distortion of the camera:
        
        
        \begin{equation*}
            K = 
            \begin{bmatrix}
                f_x & 0   & o_x \\
                0   & f_y & o_y \\
                0   & 0   & 1
            \end{bmatrix}.
        \end{equation*}
        
        In general, the camera calibration process uses images of a 3D object with a geometrical pattern (i.e., a checkerboard), which is called  \textit{calibration grid}, for matching the 3D coordinates of the pattern to the 2D image points of the same pattern. Then, the matches are used to calculate the camera parameters.
        
        For the open research challenge, we could rely only on the pitch information, because no calibration grid has been provided. For such a reason, we performed an initial setup aiming at finding patterns such as corners, intersections between the penalty areas and the goal lines, intersections between the midfield line and the sidelines, penalty area corners.
        
        We used the 2D points of those patterns and the corresponding 3D real-world coordinates of the same patterns to compute the camera matrix, the distortion coefficients, and the rotation and translation vectors. Such information has been saved to a calibration file that is used to undistort the video images. Fig. \ref{fig:comparison_calibration_frames} shows an image of the field before and after the calibration process.
        
        
        \begin{figure}[H]
            \centering
            \subfloat[\centering Image before the calibration process.]{{\includegraphics[width=10cm]{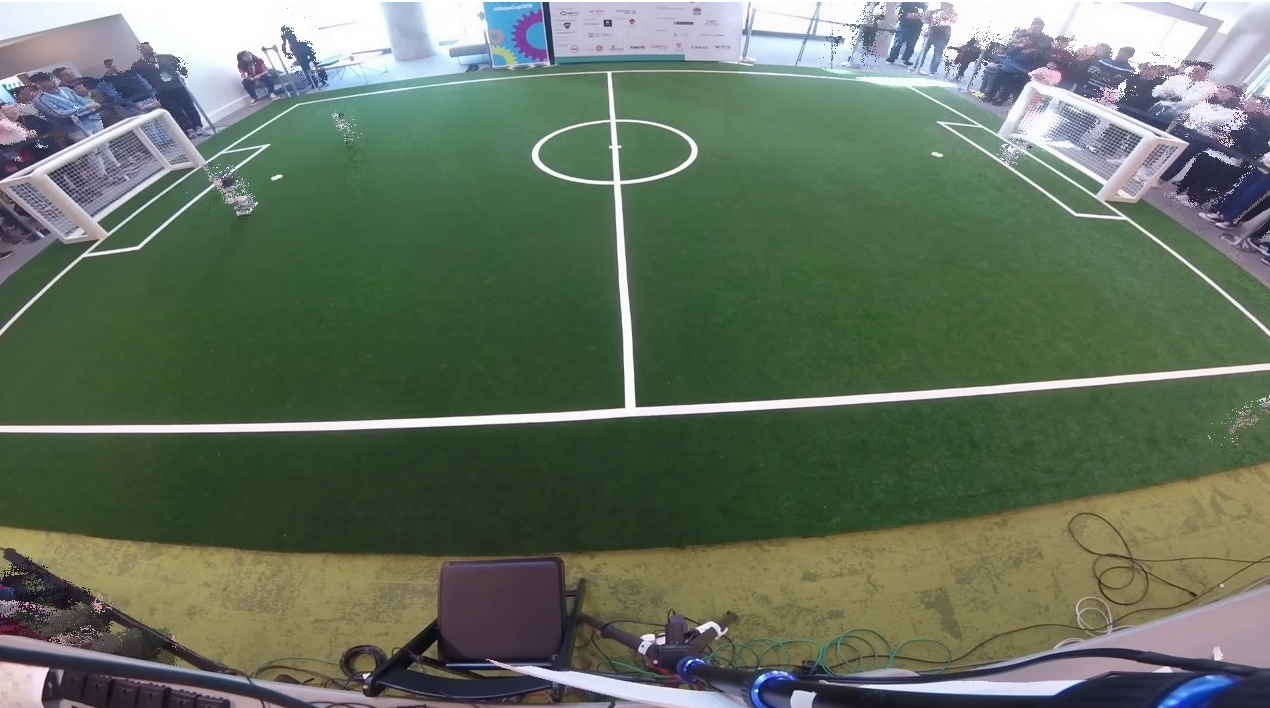} }}
            \qquad
            \subfloat[\centering Image after the calibration process.]{{\includegraphics[width=10cm]{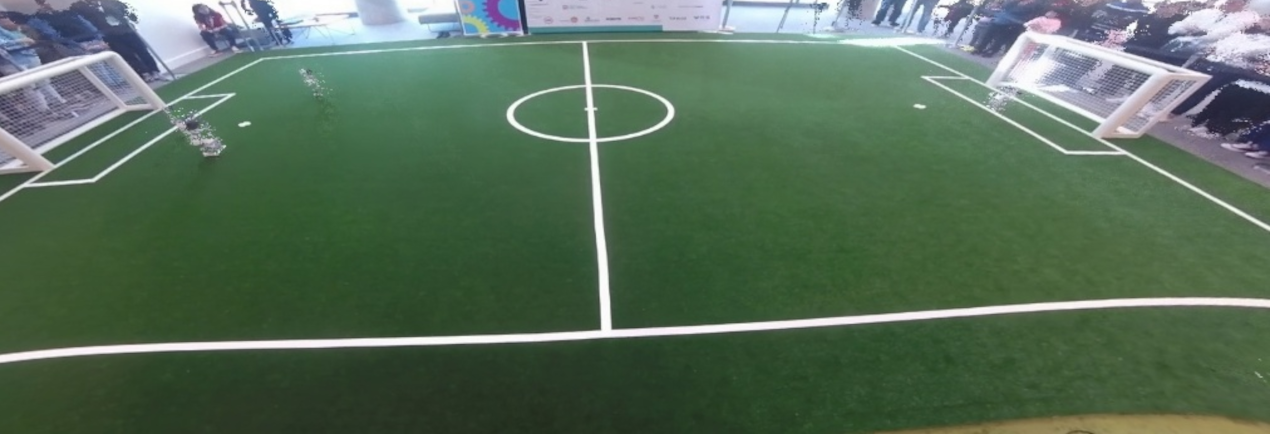} }}
            \caption{Before/after the calibration process.}
            \label{fig:comparison_calibration_frames}
        \end{figure}
        
    \subsection{Background Subtraction}
        
        \textit{Background Subtraction} (\textit{BS}) is a popular and widely used technique that represents a fundamental building block for different Computer Vision applications, ranging from automatic monitoring of public spaces to augmented reality.
        
        The BS process is carried out by comparing the current input frame with the model of the background scene and considering as foreground points the pixels that differ from the model. So, the main problem is to generate a consistent background model that is reliable with the observed scene.
        
        BS has been largely studied and many techniques have been developed for tackling the different aspects of the problem. In particular, we used a method called Independent Multimodal Background Subtraction (IMBS).
        
        \subsubsection{Independent Multimodal Background Subtraction}
        
            \textit{Independent Multimodal Background Subtraction} (IMBS) \cite{BloisiCompIMAGE2012} is a BS method designed for dealing with highly dynamic scenarios characterized by non-regular and high-frequency noise. IMBS is a per-pixel, non-recursive, and non-predictive BS method, meaning that:
        
            \begin{itemize}
                \item each pixel signal is an independent process (\textit{per-pixel});
                \item a set of input frames is analyzed to estimate the background model based on a statistical analysis of those frames (\textit{non-recursive});
                \item the order of the input frames is considered not significant (\textit{non-predictive}).
            \end{itemize}
        
            The above-listed design choices are necessary in order to achieve a very fast computation, since working at pixel level and considering each background model as independent from the previously computed ones allow us to carry out the BS process in parallel.\\
            
            \textbf{IMBS-MT}. \textit{Independent Multimodal Background Subtraction Multi-Thread} (\textit{IMBS-MT}) \cite{BloisiPRL2017} is an enhanced version of IMBS. IMBS-MT differs from the original IMBS in two aspects:
            
            \begin{enumerate}
                \item The background formation and foreground extraction processes are carried out in parallel on a disjoint set of sub-images from the original input frame;
                \item The background model is initialized incrementally, i.e., the quality of the model is increased as soon as more frame samples are available.
            \end{enumerate}
            
            IMBS-MT is designed for performing an accurate foreground extraction in real-time for full-HD images. IMBS-MT can deal with illumination changes, camera jitter, movements of small background elements, and changes in the background geometry. Fig. \ref{fig:comparison_calibration_frames} shows an example of an image processed by IMBS-MT.
            
            \begin{figure}[H]
                \centering
                \subfloat[\centering Image before the application of IMBS-MT.]{{\includegraphics[width=10cm]{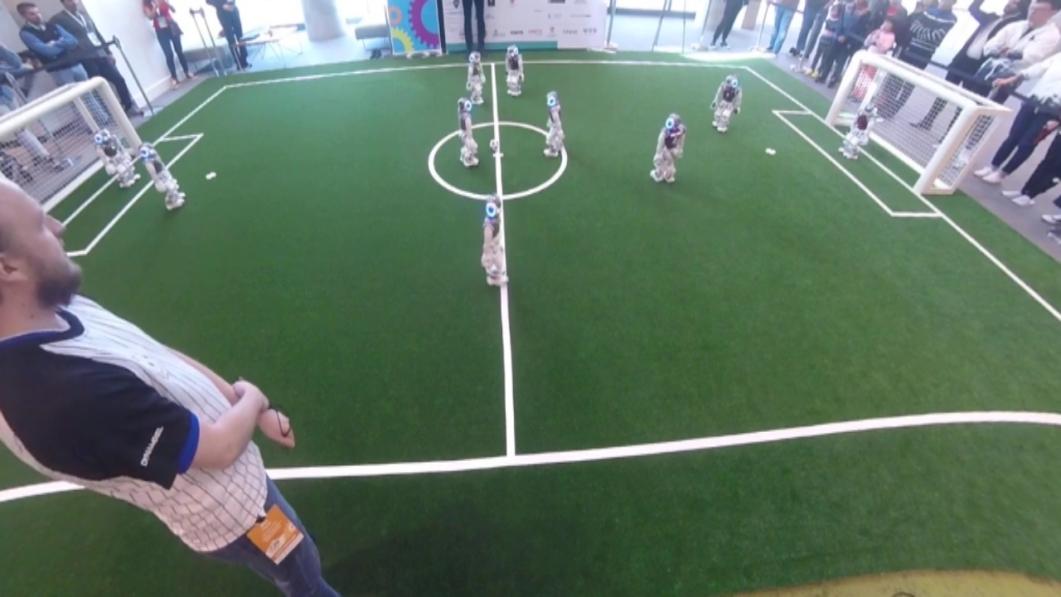} }}
                \qquad
                \subfloat[\centering Image after the application of IMBS-MT.]{{\includegraphics[width=10cm]{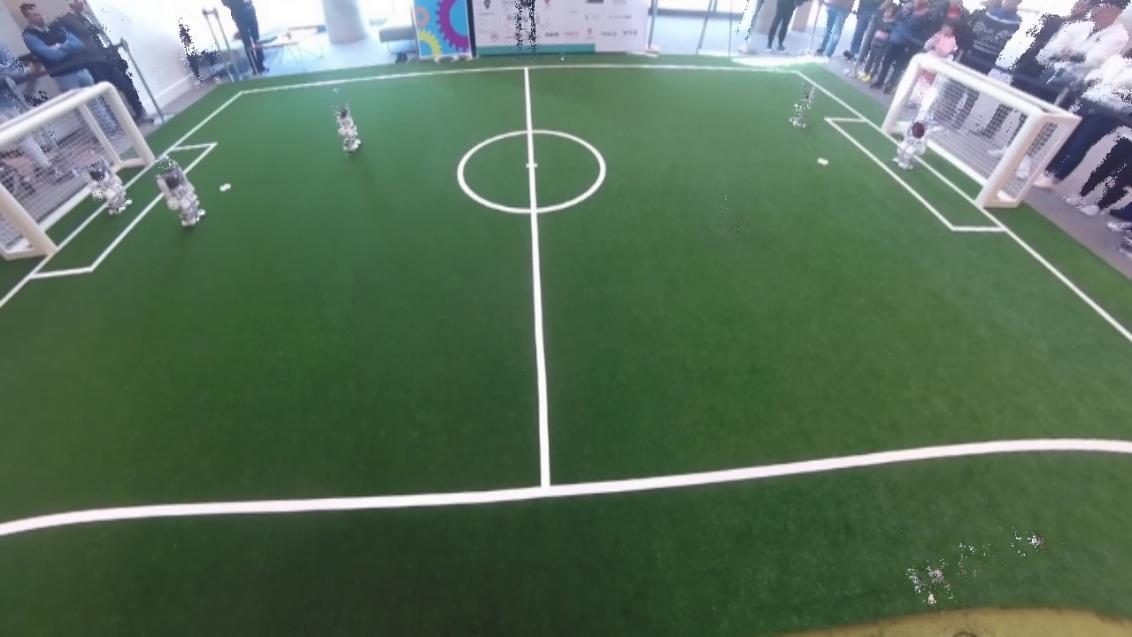} }}
                \caption{Before/after the application of IMBS-MT.}
                \label{fig:comparison_calibration_frames}
            \end{figure}
    
    \subsection{Homography}
    
    
        We calculated a homography matrix to apply a perspective transformation of the soccer field in order to obtain a bird-view of the pitch. To this end, we used an image of the soccer field as the source and a reference image of the field as the destination. In order to find the corresponding points between the two images, we trained a neural network to recognize patterns like corners, intersections between the penalty areas and the goal lines, intersections between the midfield line and the sidelines, the penalty area corners, etc.
        
\begin{figure}[H]
            \centering
            \includegraphics[scale=0.2]{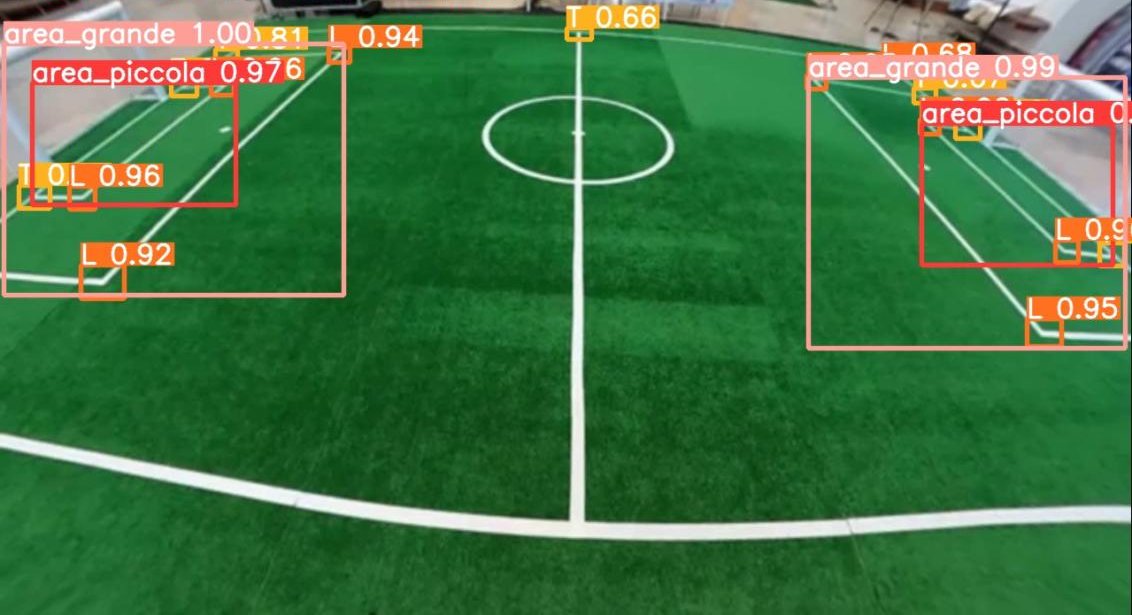}
            \caption{Patterns identified by the detector. \textit{T} labels identify intersection points and \textit{L} labels identify joining points. Other labels are used to identifies the penalty areas and the goal areas - such labels are used later to infer the field's version.}
            \label{fig:detection_t_l_penalty_area_and_goal_area}
        \end{figure}

        The network model uses \textit{YOLOv5} (a more detailed explanation of this model will be given in Section \ref{ssec:yolo}) and a proper data-set has been created for accomplishing the task. Fig. \ref{fig:detection_t_l_penalty_area_and_goal_area} shows the patterns identified by the detector.
        
        
        
        Using such a model, we can figure out the version of the pitch used in the match by analyzing if the penalty areas and the goal areas are detected. Moreover, we can calculate the set of the source points used for calculating the homography, which corresponds to the intersection and the joining points of the field lines.
        Since the set of the destination points is fixed, the homography and the pitch versions are automatically computed.
        
\begin{figure}[H]
            \centering
            \subfloat[\centering Shot captured by the camera.]{{\includegraphics[width=10cm]{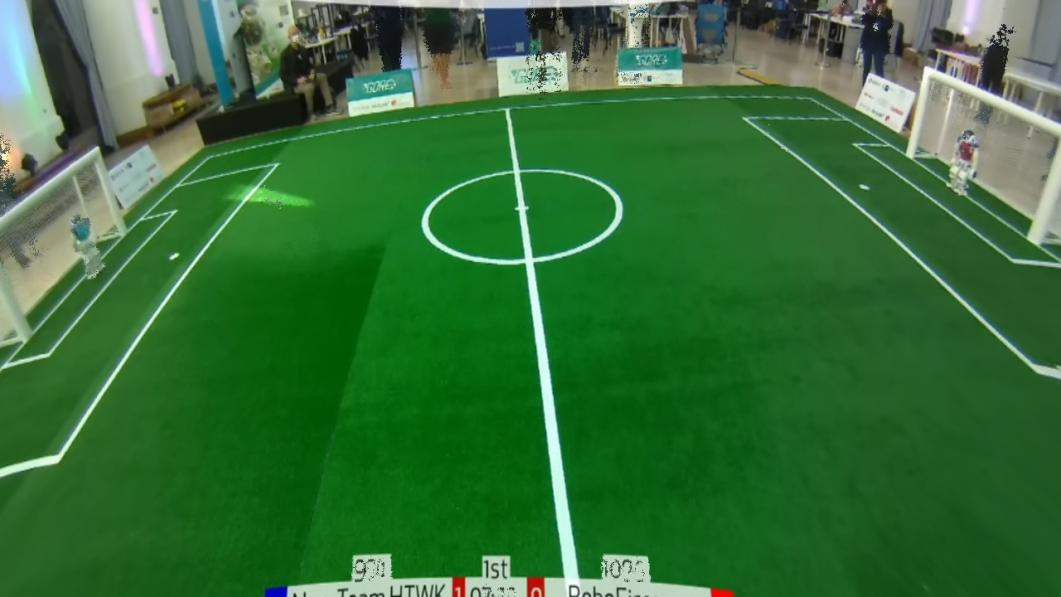} }}
            \qquad
            \subfloat[\centering Perspective transformation of the shot captured by the camera.]{{\includegraphics[width=10cm]{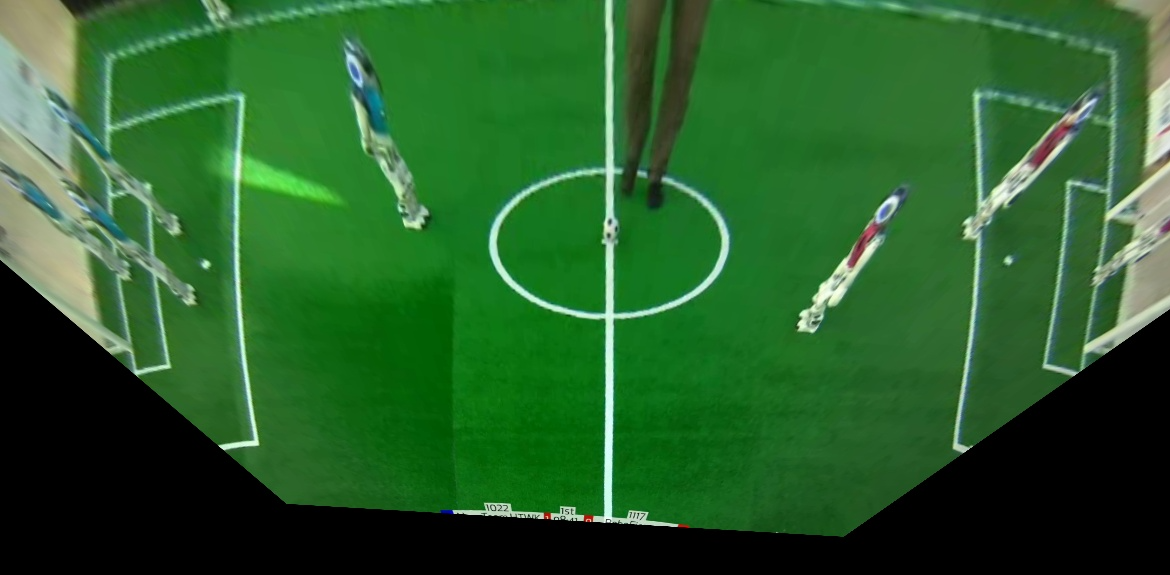} }}
            \caption{Soccer field as seen from the camera point of view and the same image after the application of the perspective transformation.}
            \label{fig:comparison_homography_frames}
        \end{figure}

        The system gives the possibility to manually choose the destination points by using the mouse. This method is usually used when the \textit{re-projection error} of the automatic homography falls above a threshold that does not guarantee a good projection result. Fig. \ref{fig:comparison_homography_frames} shows an image of the soccer field as seen from the camera's point of view and the same image after the application of the perspective transformation.
        
        
         
%% file: detection.tex
\section{Detection, Tracking, and Localization} \label{detection}
    
        \subsection{Detection} \label{ssec:yolo}
            Object detection is a technique for locating instances of objects in images or videos. In the literature, there exist many techniques for detecting objects \cite{Zaidi21} with deep learning methods. In this report, we focus on a widely used technique called YOLO.
            
            \subsubsection{YOLO}
            
            \textit{You Only Look Once} (YOLO) \cite{RedmonCoRR2015} is an object detection algorithm that is part of the one-stage detectors and uses a single network for performing both the tasks of detection and classification. It uses an end-to-end neural network on the whole image and divides the image into grid regions while predicting the rectangular boxes in each region of the grid, treating the object detection problem as a regression problem where the model only needs to perform one operation on the input.
            
            Each grid is responsible only for the targets whose centroids are within the grid, and each grid predicts the coordinates of several rectangular boxes as well as their scores. Each rectangular box corresponds to a five-dimensional output, i.e., coordinates and confidence level. Such an approach is fast in inference but less accurate.\\
            
\begin{figure}[H]
                \centering
                \subfloat[\centering Robots and ball detection.]{{\includegraphics[width=10cm]{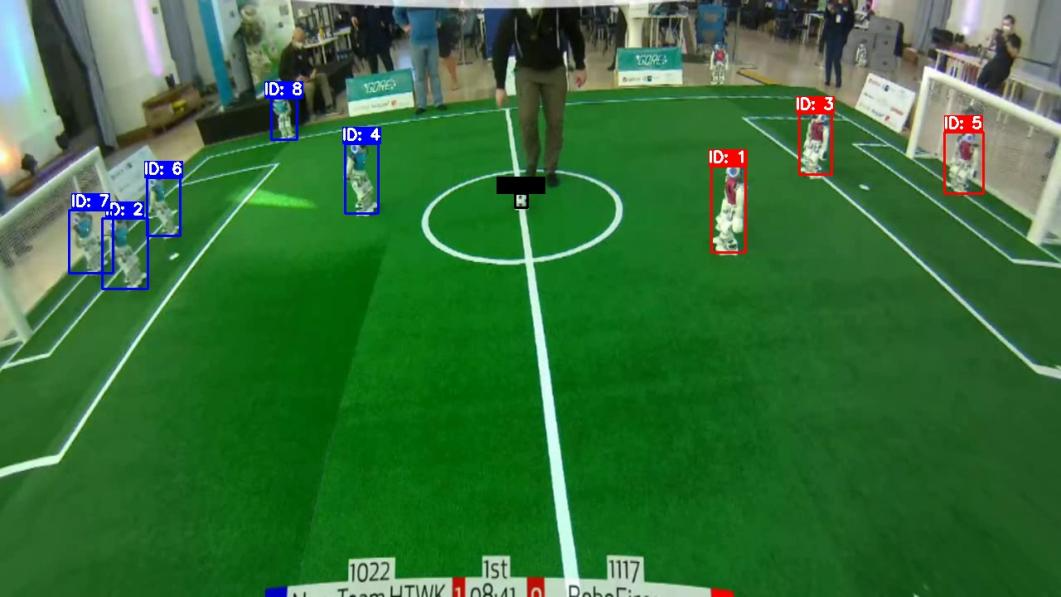} }}
                \qquad
                \subfloat[\centering Radar-like view.]{{\includegraphics[width=10cm]{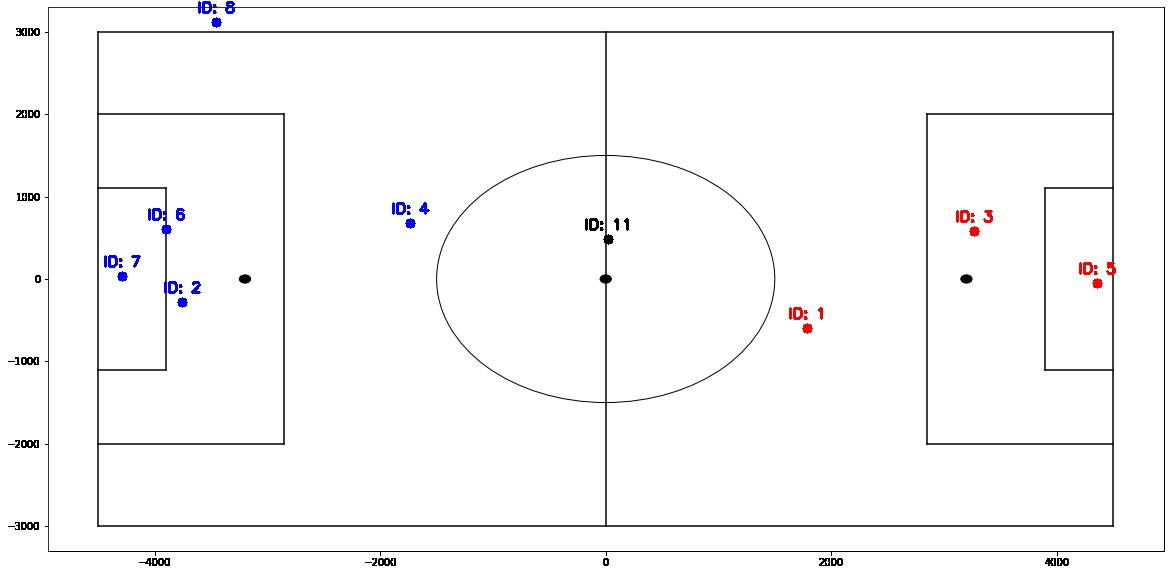} }}
                \caption{Detection in real images and perspective transformation for creating a radar-like view.}
                \label{fig:detection_and_2D_field}
            \end{figure}

            \textbf{YOLOv5}. \textit{YOLOv5} is one of the most recent version of YOLO that increases the detection speed by 50\% compared to the previous generation, i.e., YOLOv4. YOLOv5 has a model size that is only $1/10$ of that of YOLOv4. The adaptive anchor frame calculation and the use of a focused structure enhance the accuracy of the model for small target recognition. At the same time, the model has four network models with different depths, thus allowing the best balance between detection accuracy and recognition speed. Fig. \ref{fig:detection_and_2D_field} shows an image of the detection of the robots and the ball, and an image of the perspective transformation of the positions of the robots and the ball into a 2D field view.
            
     \subsection{Tracking}
        
        \textit{Tracking} is the task of predicting the positions of objects throughout a video using their spatial and temporal features. In other words, it takes the initial set of detections, assigns unique IDs, and tracks them throughout the frames of a video feed while maintaining the assigned IDs. Tracking is a two-step process that involves the use of a detection module for target localization and a motion predictor. \\
        
        The tracking systems can be classified according to the number of objects to be tracked:
        \begin{itemize}
            \item \textit{Single Object Tracking} (SOT) where the systems tracks only a single object even if there are many other objects in the same scene;
            \item \textit{Multiple Object Tracking} (MOT) where the system tracks multiple objects.
        \end{itemize}
        
        The tracker used in this report belongs to the MOT category. In particular, we used a tracker called \textit{StrongSORT} \cite{Du2022} \cite{strongsort_github}, which is an enhanced version of the popular \textit{DeepSORT} \cite{Wojke2017simple} algorithm, which in turn is an extension of \textit{SORT} \cite{Bewley2016GORU}.
        
        \subsubsection{SORT}
        
            Simple Online Realtime Tracking (SORT) is an approach that uses classical tracking techniques such as \textit{Kalman filters} \cite{Welch2020} and \textit{Hungarian algorithms} \cite{Kumar2021SS} to track objects. SORT is made of 4 key components which are as follows:
            
            \begin{enumerate}
                \item \textit{Detection}. An object detector detects the objects to be tracked. Detectors like FrRCNN, YOLO, and more are the most frequently used;
                \item \textit{Estimation}. Then, the detections are propagated from the current frame to the next one using a constant velocity model. When a detection is associated with a target, the detected bounding box is used to update the target state while the velocity components are calculated using a Kalman filter framework;
                \item \textit{Data association}. A cost matrix is computed as the Intersection-over-Union (IoU) distance between each detection and all the predicted bounding boxes from the existing targets. The assignment is solved using a Hungarian algorithm. If the IoU of the detection and the target is less than a certain threshold the assignment is rejected. This technique copes with the occlusion problem and helps in maintaining the correct IDs;
                \item \textit{Creation and Deletion of Track Identities}. This module is responsible for the creation and deletion of IDs. Unique identities are created and destroyed according to the threshold defined in the previous step. If the overlap between the detection and the target is less than a threshold the object is marked as \textit{untracked}. Tracks are terminated if they are not detected for a certain number of frames. If an old object appears again in the scene, the tracking system assigns a new ID to it.
            \end{enumerate}
            
            SORT performs very well in terms of tracking precision and accuracy, but it is affected by a high number of ID switches and failures in case of occlusion. This is due to the association matrix used for data association that is calculated only using the Euclidean distance between the detections and the tracks.
            
            DeepSORT uses a better association metric that combines both motion and appearance descriptors. The network is trained on a large-scale person re-identification data-set, making it suitable for tracking context. The association metric model used in DeepSort is based on the cosine distance metric \cite{Wojke2018deep}, which is suitable for calculating the similarity between two feature vectors.
            
            In this work, we used an enhanced version of DeepSORT: StrongSORT, which improves the standard DeepSORT in the following points: 
            
            \begin{itemize}
                \item the use of a stronger appearance feature extractor in place of the simple CNN used by DeepSORT. It uses a ResNet50 model as a backbone network that can extract much more discriminative features;
                \item the use of a feature updating strategy that involves the exponential moving average (EMA). The EMA updating strategy enhances the matching quality and reduces time consumption;
                \item the use of a more robust Kalman filter in place of the vanilla one implemented in DeepSORT.
            \end{itemize}
            

%% file: pose.tex
\section{Pose Estimation and Fall Detection}\label{pose}

    \subsection{Pose Estimation}
        
        Multi-person pose estimation is an important task that can be used in different domains, such as action recognition, motion capture, sports, etc. The task predicts a pose skeleton for every person in an image.
        
        The skeleton consists of key points, or joints, that identifies specific body parts such as ankles, knees, hips, and elbows. Multi-person pose estimation problem can be approached in two ways:
        \begin{enumerate}
            \item \textit{top-down}, which first applies a person detector and then runs a pose estimation algorithm for every detected person;
            \item \textit{bottom-up}, where all the key-points are detected in a given image and then grouped by the entity instances.
        \end{enumerate}
        
        The bottom-up approach is usually faster than the top-down one since it finds key points once and does not rerun pose estimation for each person.
        

        \subsubsection{OpenPose}
        
            \textit{OpenPose} \cite{CaoIEEETPAMI2019} is the first real-time multi-person system to jointly detect the human body, feet, hands, and facial key points on a single image. OpenPose uses a CNN model such as VGG-19 for feature map extraction that then is used for a multi-stage CNN pipeline to generate the Part Confidence Maps (PCM) and the Part Affinity Fields (PAF). In the last step, the PCM and the PAF are processed by a greedy bipartite matching algorithm to obtain the poses for each entity in the image.\\
            
            \textbf{Lightweight OpenPose.} In our system, we used a slightly optimized version of OpenPose, called \textit{Lightweight OpenPose} \cite{Osokin2018}. Such a method allows real-time inference on CPU with a negligible accuracy drop. Some of the improvements achieved by the Lightweight OpenPose method are described in the following points:
            
            \begin{itemize}
                \item \textit{Lightweight Backbone}. A lighter but still effective network is used. The commonly used network is MobileNet V1 \cite{Howard17};
                \item \textit{Lightweight Refinement Stage}. To produce new estimations of PCMs and PAFs, the refinement stage takes features from the backbone network concatenated with the previous estimation of PCMs and PAFs. In order to share the computation between PCMs and PAFs and realize a consistent speed-up, a single prediction branch is used in the initial and refinement stages. The last two layers are not shared between PCMs and PAFs and produce the model output;
                \item \textit{Fast Post-processing}. Key points extraction is performed in a parallel way.
            \end{itemize}
            
\begin{figure}[h!]
                \centering
                \includegraphics[scale=0.25]{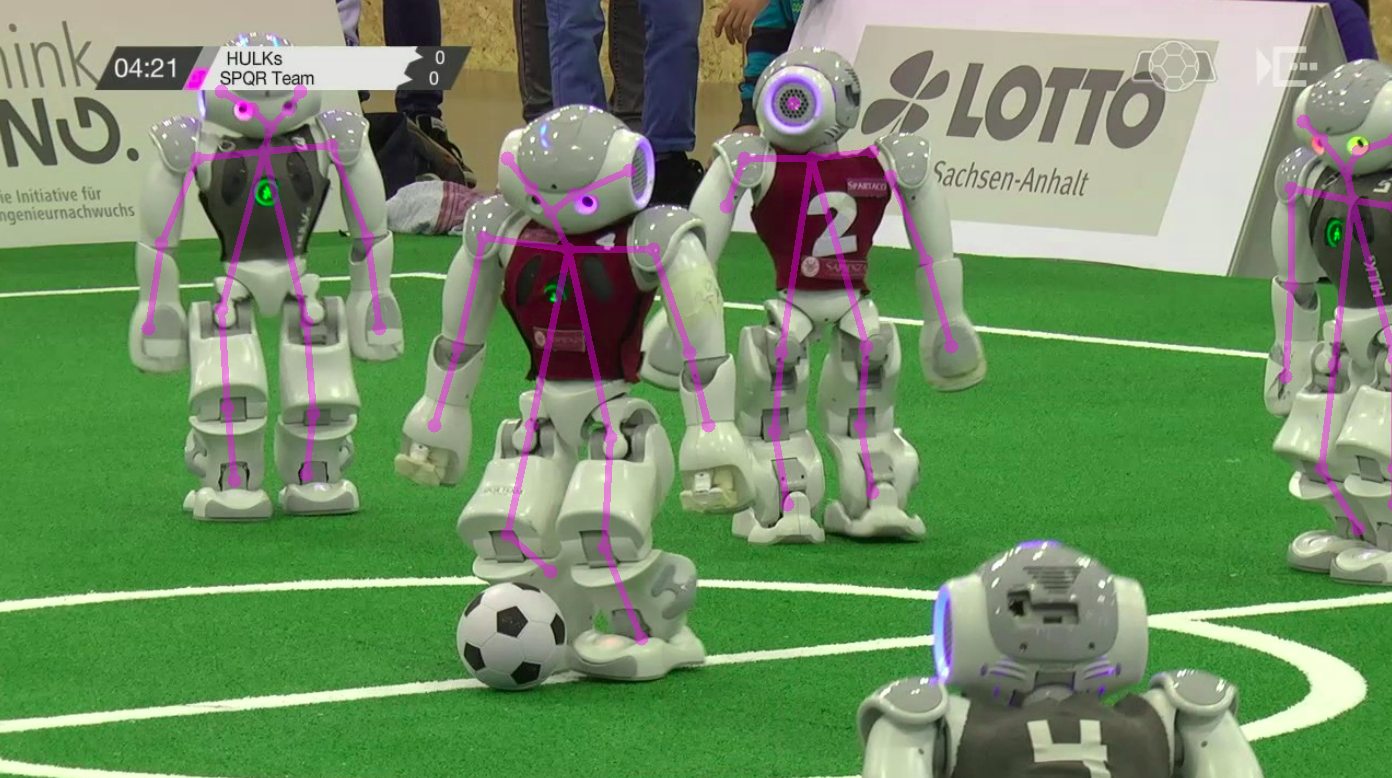}
                \caption{Pose estimation for NAO robots.}
                \label{fig:nao_pose_estimation}
            \end{figure}

            Fig. \ref{fig:nao_pose_estimation} shows an example of pose estimation performed on a single frame of a RoboCup soccer game.
            
            In the next subsection, we will explain how we created the training data set and how the Lightweight OpenPose method has been adapted to the NAO robots.
            
    \subsection{Fall Detection}
    
        Human action recognition has become an active research area in recent years, as it plays a significant role in video understanding. In general, there exist many methods for understanding human actions, and one of these involves the use of dynamic skeletal information.
        
        The dynamic skeletal information can be represented by a time series of human joint locations, in the form of 2D or 3D coordinates. Human actions can then be recognized by analyzing the motion patterns of such coordinates.
        
        The skeleton and the joint trajectories of the human bodies are robust to illumination changes and scene variations, and they are easy to obtain using depth sensors or 3D pose estimation algorithms.
        
        In general, the approaches can be categorized into:

        \begin{itemize}
            \item \textit{Feature Based methods}, that design several handcrafted features to capture the dynamics of the joint motion. The most common features are covariance matrices of joint trajectories, relative positions of joints, and rotations and translations between body parts;
            \item \textit{Deep Learning methods}, that involve the use of recurrent neural networks (RNN) and temporal CNNs.
        \end{itemize}

        By using the skeletal information retrieved through the Lightweight OpenPose method and observing the temporal dynamic of a single skeleton, it is possible to infer the type of action that is being performed by the robots. In particular, we are interested in capturing the fall-down action of the robots.
    
        \subsubsection{ST-GCN}
        
\begin{figure}[H]
                \centering
                \includegraphics[scale=0.275]{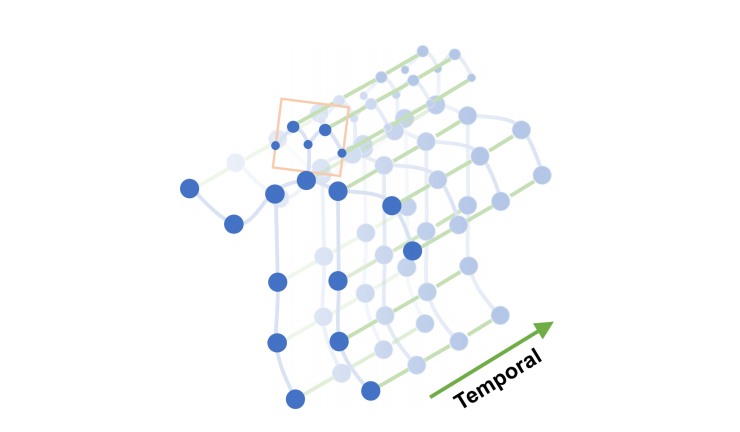}
                \caption{Spatio-Temporal graph of a skeleton sequence. Blue dots denote the body joints and blue edges between body joints define natural connections in human bodies. [\href{https://arxiv.org/pdf/1801.07455.pdf}{Image Source}]}
                \label{fig:stgcn_skeleton}
            \end{figure}

            The \textit{Spatial-Temporal Graph Convolutional Network} (ST-GCN) \cite{YanCoRR2018} is a \textit{graph convolutional network} (\textit{GCN}) able to estimate actions from the spatio-temporal graph of a skeleton sequence. An example of spatio-temporal graph is shown in Fig. \ref{fig:stgcn_skeleton}.
            
            There are two ways to perform convolutions on graphs. The first approach performs convolutions in the spatial domain, while the second one is in the spectral domain.
            
            The ST-GCN method uses convolution in the spatial domain, where the data is represented in the form of graph nodes and connections between them. Since we are dealing with graphs, the standard convolution cannot be applied to them, so, we use a  modified convolutional kernel that only operates on the direct neighbors of the node.
            
            Moreover, it divides the neighbor set into several subsets and applies different weights to these subsets during the convolution. ST-GCN proposes three partition strategies to divide the neighbor set:
            
            \begin{enumerate}
                \item \textit{Uni-labeling}. All nodes in a neighborhood are in the same subset;
                \item \textit{Distance partitioning}. The root node is inserted in a subset (distance 0) and the remaining neighbors into another subset (distance 1).
                \item \textit{Spatial configuration}. The nodes are partitioned according to their distances to the skeleton gravity center and compared with the root node. Three subsets are then created: one subset for the root node; one subset for the centripetal nodes, which have a shorter distance than the root node; and one subset for the centrifugal nodes, which have a longer distance than the root node.
            \end{enumerate}
            
\begin{figure}[H]
                \centering
                \includegraphics[scale=0.3]{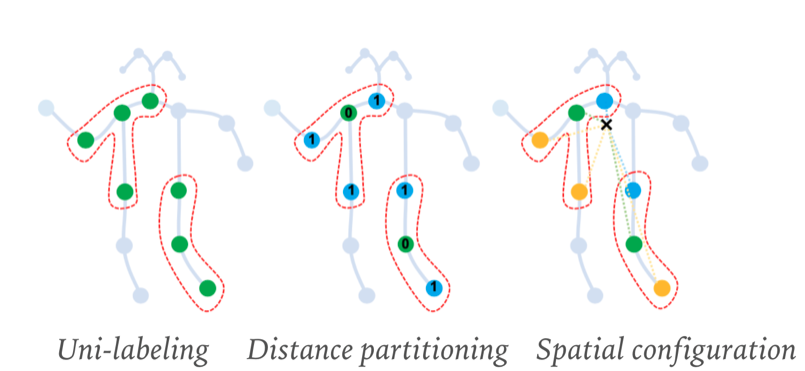}
                \caption{Partitioning methods. [\href{https://arxiv.org/pdf/1801.07455.pdf}{Image Source}]}
                \label{fig:stgcn_convolution_partitioning}
            \end{figure}

            Fig. \ref{fig:stgcn_convolution_partitioning} shows the partitioning methods above described.
            
\begin{figure}[H]
                \centering
                \includegraphics[scale=0.3]{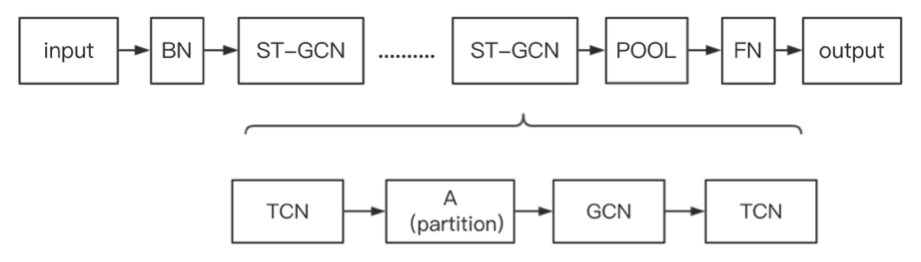}
                \caption{ST-GCN architectural scheme.}
                \label{fig:stgcn_arch}
            \end{figure}

            The network contains several spatio-temporal convolution blocks. Each of these blocks performs three actions:
            (1) temporal convolution, partition, (2) graph convolution and, in order to optimize the results, (3) a second temporal convolution. Fig. \ref{fig:stgcn_arch} shows the ST-GCN architectural scheme. 
            
            In this work, the ST-GCN method is used to identify the fallen robots. Due to the low-resolution images, the robot skeleton may be missing some of its limbs. To solve this issue, an additional fallback method has been implemented to detect falls. In particular, the shape of the bounding box obtained in the robot detection phase is used: if the aspect ratio of the bounding is below a specific threshold, the robot is considered is marked as "fallen".

%% file: illegal.tex
\section{Illegal Defender} \label{illegal}
    
    The tracking results are used to check if the \textit{illegal defender} foul is detected.
    An illegal defender context happens when three players from the same team are in the same penalty area. To be more specific, the \textit{illegal defender} works as follows: the positions of the robots in the field are extracted from the tracking data, then, the robots of each team are identified and their positions are taken into account.
    
\begin{figure}[H]
        \centering
        \subfloat[\centering No illegal defender scenario.]{{\includegraphics[width=10cm]{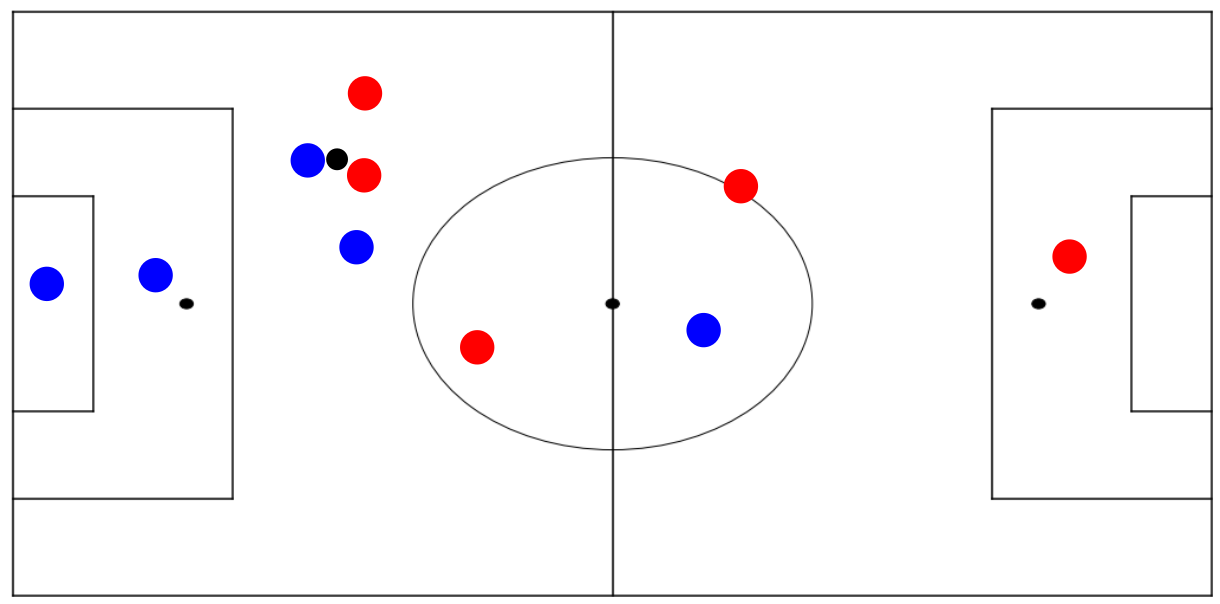}}}
        \qquad
        \subfloat[\centering Illegal defender scenario. Purple square identifies the group of players involved in a foul.]{{\includegraphics[width=10cm]{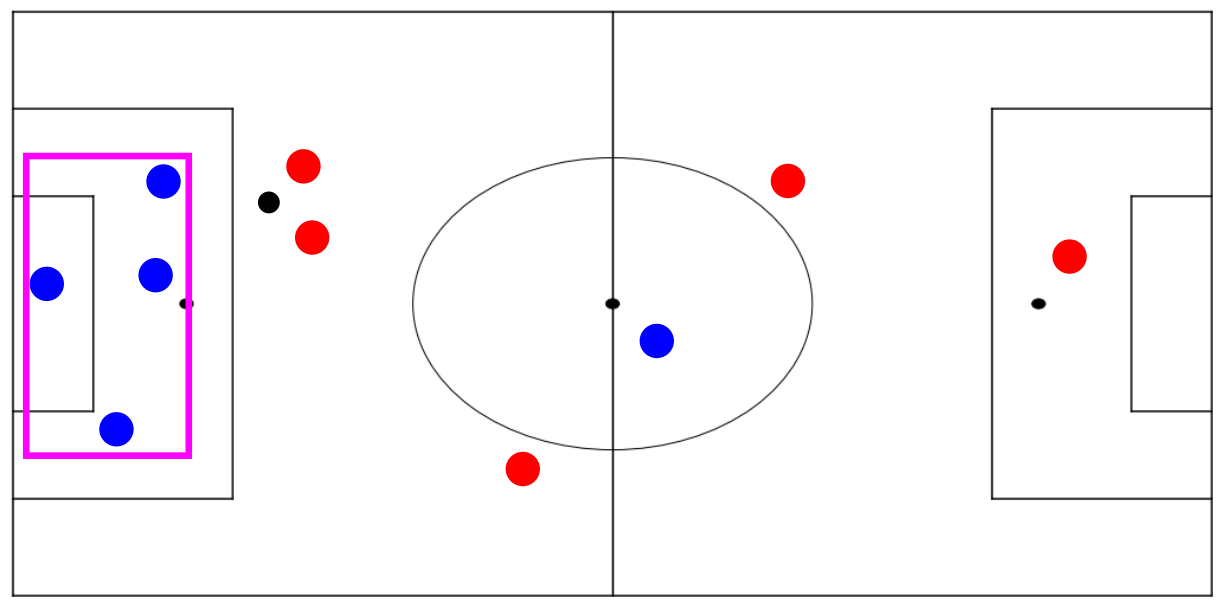}}}
        \caption{Visual difference between an illegal defender scenario and a no illegal defender scenario.}
        \label{fig:illegal_defender_and_no}
    \end{figure}

    If more than 3 robots from the same team are in the range of coordinates representing the penalty area, then an \textit{illegal defender} foul is detected and a counter is increased by 1. In the end, the total number of illegal defender fouls per team is returned. Fig. \ref{fig:illegal_defender_and_no} shows two images that visually explain when an illegal defender foul occurs and when not.
    
%% file: stats.tex
\section{Data Association and Statistics} \label{stats}

    \subsection{Data Association}
    
        \textit{Data association} consists of performing the fusion between multiple homogeneous or heterogeneous sources in order to have a unique reference system from which extracting global information about the monitored environment.
        
        In our case, it consists in aligning the player tracking data with the data coming from the \textit{GameController}. Such a step improves the accuracy of the available data and obtains additional information about the players, such as the team membership, the jersey number, whether the robot is leaving the field or not, whether the robot is dropped or inactive, and whether it committed a foul, etc. \\
        
        In our system, the data association process occurs in the first few frames of the game. The robot positions extracted from the tracking system and the positions from the \textit{GameController} are associated using a K-Means approach. In some cases, such an association process turns out to be inaccurate due to the imprecision of the position data in the \textit{GameController}. In fact, such positions are calculated by the robots themselves during the game, hence with a limited perspective.
    
    \subsection{Statistics}
    
        Match statistics and analysis computation are one of the long-term goals of the Open Research Challenge. The statistics module focuses at estimating the \textbf{heatmaps} and the \textbf{trackmaps} of robots and ball, \textbf{pass and shot maps}, and \textbf{ball possession}. All the statistics have been computed using the \textit{game\_data} file converted into a data frame on which we carry out the data analysis operation filtering to obtain new data frames. 

    \subsubsection{Heatmaps of robots and ball}
        A heatmap shows the most occupied locations by each robot and ball in the 2D plan representing the pitch top view. The system allows one to select a specific robot by its id or to choose the ball (ID 11). A new data frame called \textbf{player\_data} is created from the main one by filtering all the rows with different ids from the chosen one. Then, all the data about the player or the ball is plotted on a graph whose density depends on the most occupied area by the chosen robot or ball.  
    
        \begin{figure}[H]
            \centering
            \includegraphics[scale=0.45]{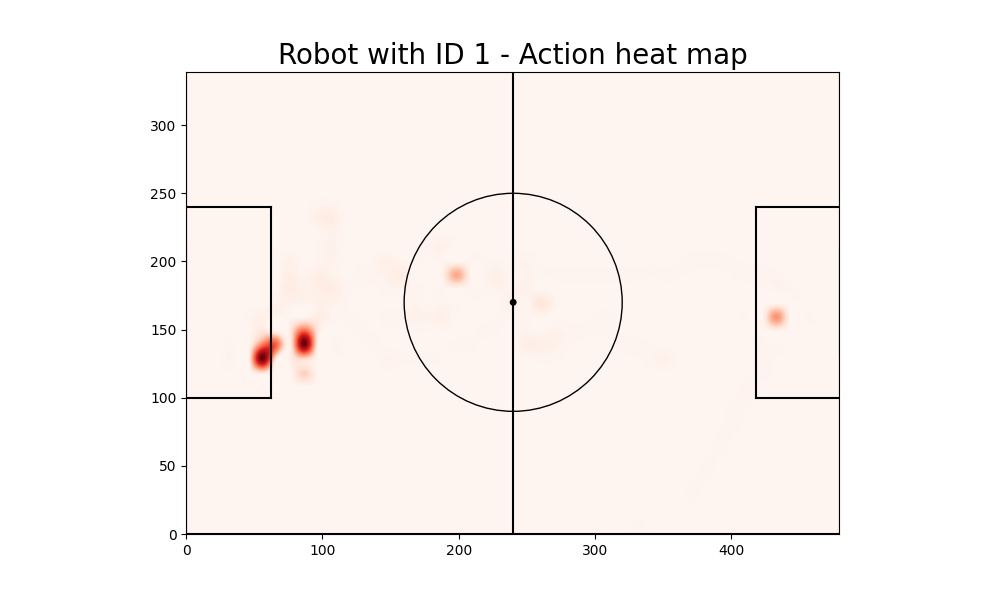}
            \caption{Heatmap of the robot with ID 1.}
            \label{fig:heatmap}
        \end{figure}
    
    \subsubsection{Trackmaps of robots and ball}
        A track map shows all the points touched by each robot and ball in the 2D plan. It is estimated in a similar way to the heatmap, but instead of using a density graph, all the positions $$(x, y)$$ of the chosen robot or ball are projected on the 2D field model.
            
        \begin{figure}[H]
            \centering
            \includegraphics[scale=0.4]{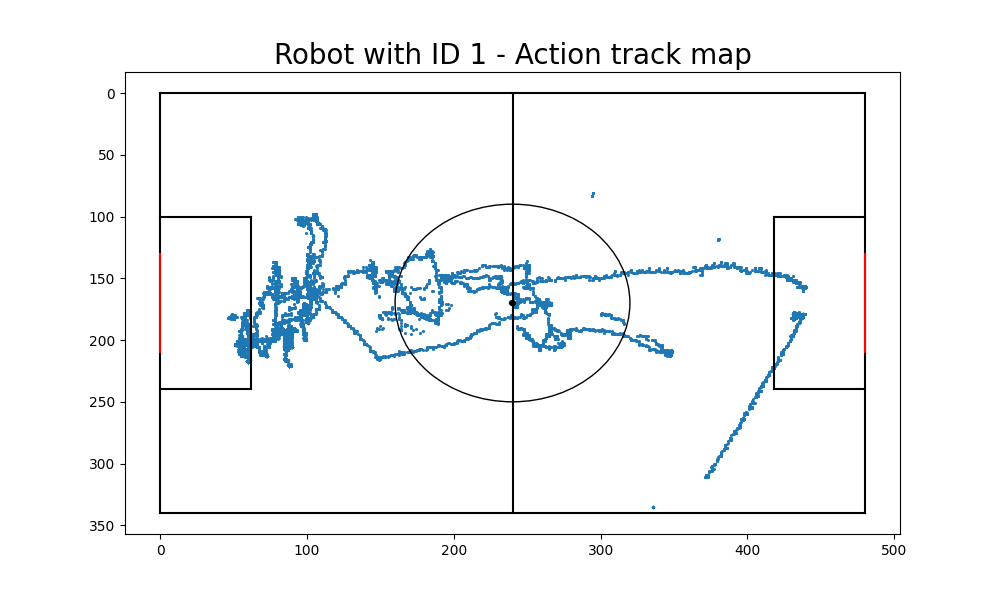}
            \caption{Trackmap of robot with ID 1.}
            \label{fig:trackmap}
        \end{figure}
            
    \subsubsection{Pass and shot map} 
        Pass and shot map show all the passes, shots, shots on target, and goals of each team in different colors. A new filtered data frame called \textit{ball} is obtained from the main one and includes all the information about the ball. Then, other 2 data frames are created from \textit{ball}: the first one called \textit{ball0} stores the information about ball at the current frame $n$, while the second one, \textit{ball1} stores information about the ball at frame $n + 5$. Then, the euclidean distance between \textit{ball0} and \textit{ball1} is calculated for extracting the number of passages. 
        
        If this distance is greater than $50$ and less than $70$ a pass counter is increased. If the ball position at frame $n$ is outside the area and it is inside at frame $n+5$, a shot counter is increased depending on which area is located.
        
        In order to calculate the shots on target, we consider only the y-coordinates related to the goal area.
        
        To calculate the goals, if the ball position at frame $n$ is outside the goal and  goes over it at frame $n + 5$, a goal count is increased by 1.     
            
        \begin{figure}[H]
            \centering
            \includegraphics[scale=0.4]{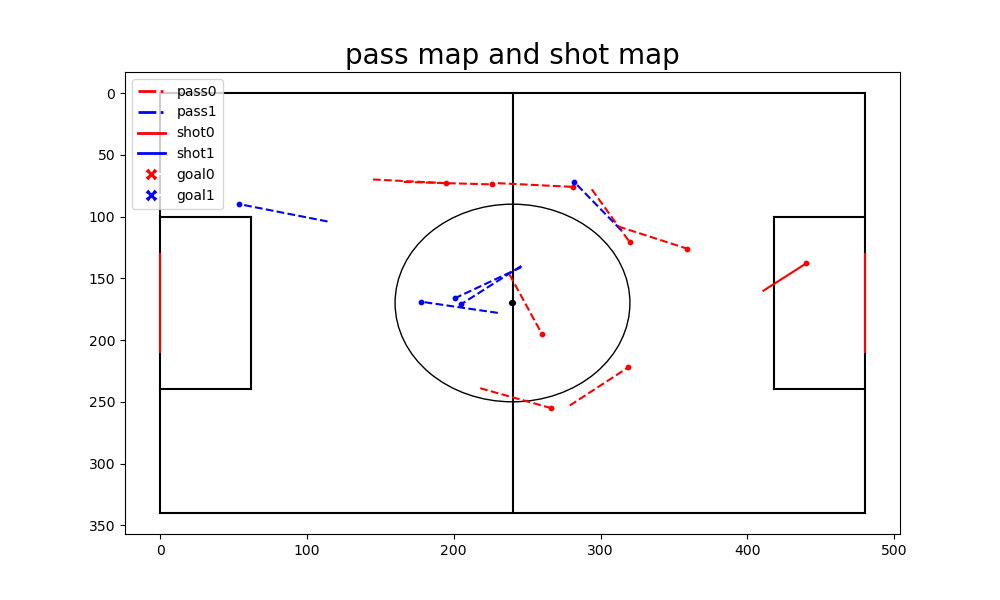}
            \caption{Map of pass and shot.}
            \label{fig:passmap}
        \end{figure}
            
    \subsubsection{Ball possession} 
        This section calculates the numerical values for the ball possession of each team. Three filtered data frames are created from the main one, namely: \textit{team0}, \textit{team1}, and \textit{ball}.
        
        For each frame in which the ball is detected and tracked, the euclidean distance between the ball position and \textit{team1} and \textit{team0} is calculated. If the distance between the ball and \textit{team0} is less than the distance between the ball and \textit{team1}, the possession counter for team 0 is increased, otherwise, we increase the possession counter for team 1.
        
        At the end of the game, the percentage for the ball possession for both teams is calculated.

%% file: gui.tex
\section{GUI} \label{gui}
    A graphical user interface (GUI) has been implemented using the Python graphic module \textit{tkinter}. From the GUI it is possible to control all the MARIO flow of execution, from the calibration process to the calculation of the final stats.
    
    The GUI consists of 3 windows (\textit{Configuration, Tracking, and Analysis}) accessible using buttons.

    \subsection{Configuration}
        The main window, called \textit{MARIO} is the configuration window where you can choose the video of the match to analyze, the extrinsic parameters (optional), the game controller (optional), and the calibration file (if your video is not calibrated).
        Then, you can run the calibration and the background subtraction modules or directly go to the tracking step.
        
        In addition, there are three icon buttons which link to
        \href{https://sites.google.com/unibas.it/wolves}{UNIBAS WOLVES}, \href{http://spqr.diag.uniroma1.it/}{SPQR team} and \href{https://portale.unibas.it/site/home.html}{UNIBAS} web pages. 

            \begin{figure}[!h]
                \centering
                \includegraphics[scale=0.7]{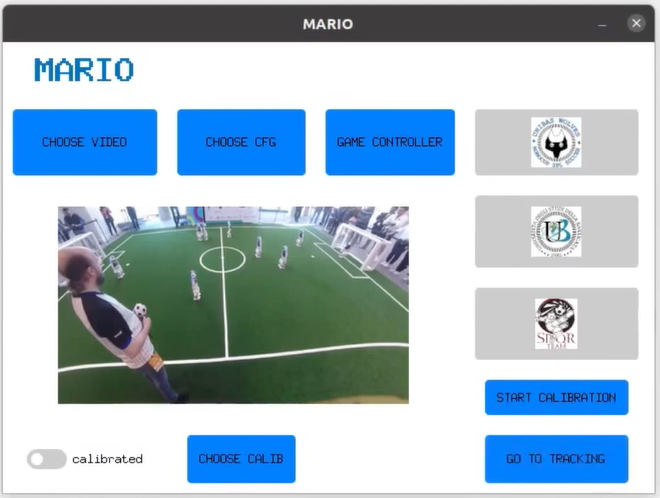}
                \caption{Main window of MARIO application.}
                \label{fig:mario-main}
            \end{figure}

    \subsection{Tracking}
        After the calibration phase that outputs a calibrated video, a second window called \textit{MARIO-Tracking} can be used to evaluate the Homography. Homography is computer before starting the tracking. In this window, 2 videos are shown: the first one is the tracking video with bounding boxes and poses estimation (only for robots) for each robot, and the second one is the tracking on the 2D model of the pitch.
        
        At the end of this phase, a file (game\_data) is created (in csv format) containing all the data about the position, the team membership, the jersey number, the ID, and the frame of each robot.
    
            \begin{figure}[!h]
                \centering
                \includegraphics[scale=0.7]{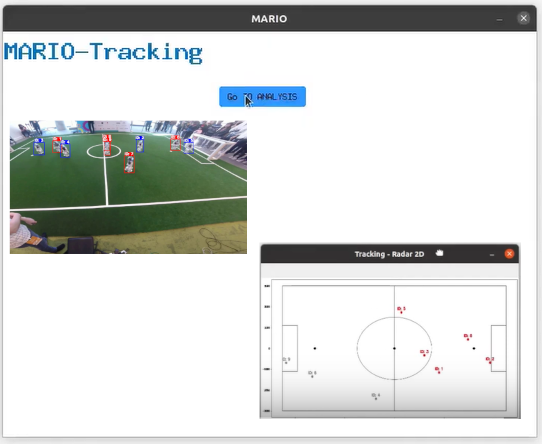}
                \caption{Window for displaying tracks.}
                \label{fig:mario-tracking}
            \end{figure}

    \subsection{Analysis}
        You can subsequently open the analysis window by pressing the \textit{GO TO ANALYSIS} button. In this third window, called \textit{MARIO-Analysis}, you can estimate the robot heatmaps and track maps, pass and shot maps, and calculate the numerical stats of the match,  such as goals, ball possession, total attempts, attempts on target and total passes that are shown in a window with a soccer match scoreboard design.
    
            \begin{figure}[H]
                \centering
                \includegraphics[scale=0.7]{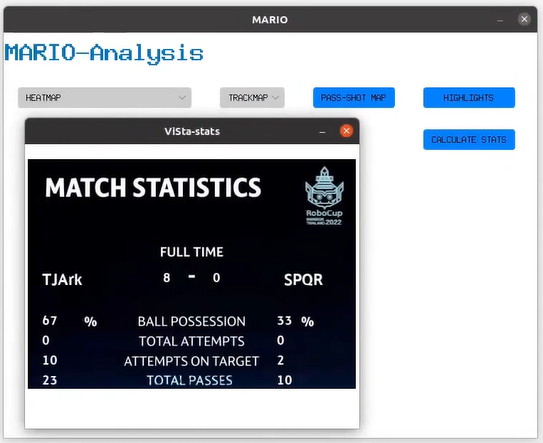}
                \caption{Window for displaying analysis.}
                \label{fig:mario-analysis}
            \end{figure}

%% file: dataset.tex
\section{Data-sets} \label{datasets}
In this section, we present two data-sets used for training the networks for accomplishing the tasks of detection and pose estimation. In particular, the pose estimation data-set is a novelty in the scientific community and it is publicly available.

\subsection{Detection data-set}\label{ssec:yolodataset}

    To carry out robots and ball detection, the database of images provided by the RoboCup SPL has been used. Each team provided $5000$ images captured from the matches of RoboCup SPL 2019, and for each one, they carried out the labeling as explained in the rule book \cite{RoboCupTechnicalCommittee2022}. The data set consists of 35000 images, and we used 24000 images as a training set, 3000 images as a validation set, and the remaining 8000 images as a testing set. \\
    
    The data-set used for the detection is distributed as-is at the following  \href{https://drive.google.com/file/d/1mqb2HHyt_3aX1dnhJMlEUQFwKH8yXxjm/view?usp=sharing}{link}.

\subsection{UNIBAS NAO Pose data-set}
        
    NAO robots share the same body structure as human beings. However, from a model point of view, the PCMs and PAFs calculated by this method show the existence of substantial differences between robots and humans. For such a reason, we built a new data set called \textit{UNIBAS NAO Pose Data-set}. \\
    
    To create the data set, we used the COCO Annotator tool \cite{coco_annotator}, a web-based image annotation tool designed for efficiently labeling images for image localization and object detection. All annotations share the same basic data structure. The pose is made of up to 18 key points: ears, eyes, nose, neck, shoulders, elbows, wrists, hips, knees, and ankles. The annotations are stored using a JSON structure. \\
    
    The data-set is distributed as-is at the following \href{https://drive.google.com/drive/folders/1L_JNu6AYqcQWZVo-Q8nVVosUBb-DjXrV}{link}.

%% file: results.tex
\section{Experimental Results} \label{results}
    
    A video which highlights the main features of MARIO is available at the following \href{https://www.youtube.com/watch?v=eutyWaQ4-oU}{link}.

    \subsection{YOLO Results}
        
            To train YOLOv5, the data-set described in Section \ref{ssec:yolodataset} has been converted into the YOLO format. Then, YOLOv5 has been trained for 270 epochs obtaining the following results:
        
            \begin{itemize}
                \item \textit{Precision}: 0.982;
                \item \textit{Recall}: 0.932;
                \item \textit{Mean Average Precision (mAP) 0.5}: 0.953;
                \item \textit{Mean Average Precision (mAP) 0.5:0.95}: 0.822.
            \end{itemize}
            
\begin{figure}[!h]
                \centering
                \subfloat[\centering Precision.]{{\includegraphics[width=5cm]{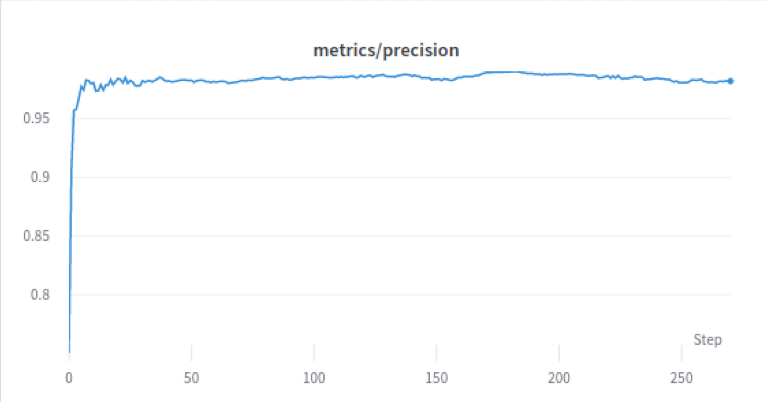} }}
                \qquad
                \subfloat[\centering Recall.]{{\includegraphics[width=5cm]{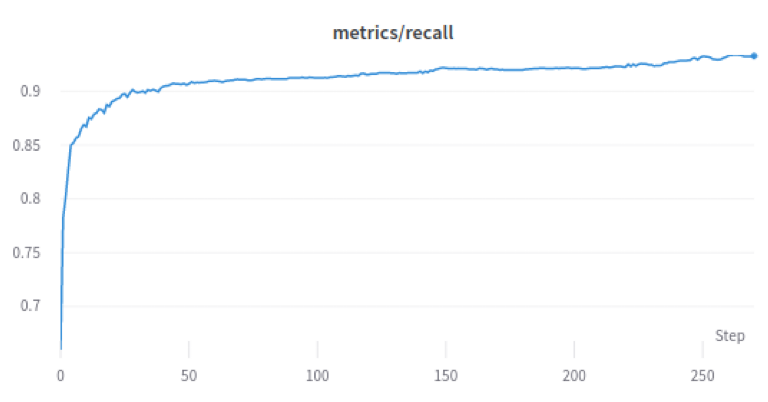} }}
                \qquad
                \subfloat[\centering mAP 0.5.]{{\includegraphics[width=5cm]{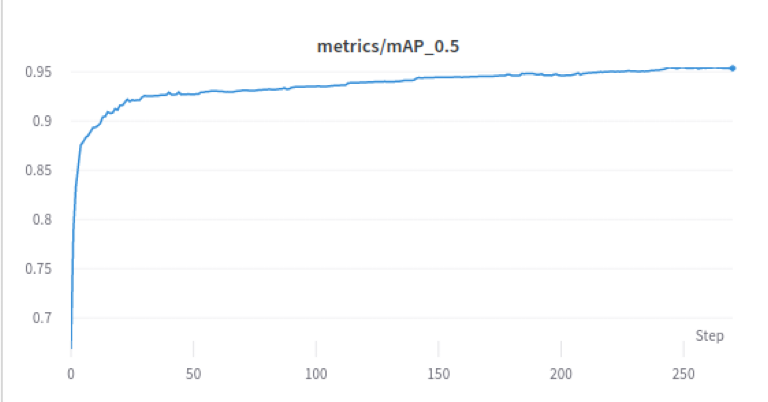} }}
                \qquad
                \subfloat[\centering mAP 0.5:0.95.]{{\includegraphics[width=5cm]{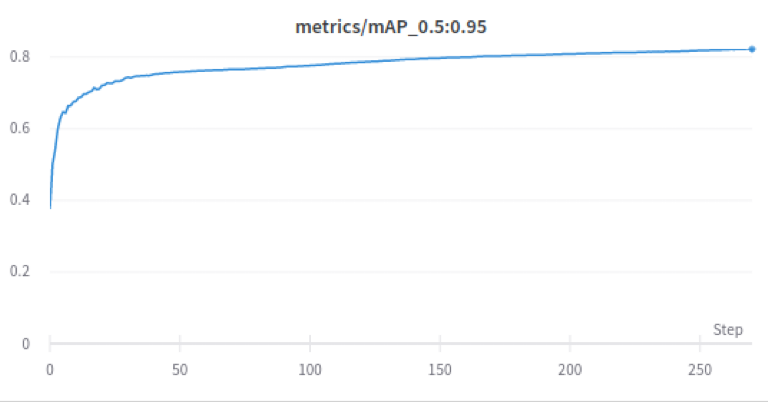} }}
                \caption{Metrics of the YOLOv5 model.}
                \label{fig:precision_recall_map}
            \end{figure}

            Figure \ref{fig:precision_recall_map} shows all the metrics scored by the YOLOv5 model used.
            
    \subsection{Run-time Performance}
    
        Run-time performance is evaluated on three different types of hardware:
        
        \begin{itemize}
            \item a \textit{low-end hardware} with CPU Intel i3-6006U, GPU Intel Graphics HD 520, and 8 GB of RAM;
            \item a \textit{middle-end hardware} with CPU Ryzen 7 5700U, GPU NVidia RTX 3050 Mobile with 4 GB of vRAM, and 16 GB of RAM;
            \item an \textit{high-end hardware} with CPU Intel Xeon W-2135, GPU Nvidia Quadro P4000 with 8 GB of vRAM, and 32 GB of RAM.
        \end{itemize}
        
        The performance has been evaluated with and without the pose estimation system. Pose estimation is a very computationally expensive operation and it significantly degrades the performance of the system.
        
        Table \ref{table:runtime_performance} shows the performance of MARIO, in terms of average frames per second, with all the tested hardware scenarios, with and without pose estimation.
        
        \begin{table}[t]
            \centering
            \caption{Performance of MARIO - in terms of average frames per second - with all the tested hardware scenarios, with and without pose estimation.}
            \label{table:runtime_performance}
            \begin{tabular}{|c|c|c|} 
                \hline
                                       & \textbf{w/ Pose Estimation} & \textbf{w/o Pose Estimation}  \\ \hline
                \textbf{Low-End HW}    &                           3 &                            7  \\ \hline
                \textbf{Middle-End HW} &                           6 &                           16  \\ \hline
                \textbf{High-End HW}   &                          10 &                           20  \\ \hline 
            \end{tabular}
            
        \end{table}

%% file: conclusions.tex
\section{Conclusions} \label{conclusions}

    In this technical report, we described MARIO, modular end-to-end architecture for computing visual statistics in RoboCup SPL. Each module of MARIO performs a specific task, namely:
    
    \begin{itemize}
        \item camera calibration;
        \item background subtraction;
        \item homography;
        \item tracking and localization;
        \item pose estimation;
        \item fall detection;
        \item illegal defender;
        \item data association and statistics.
    \end{itemize}
    
    All these modules are wrapped around a simple and intuitive graphical user interface (GUI). \\
    
    The system includes many novelties such as automatic homography computation, the pose estimation task, and the creation of a custom data set. From the experimental results, MARIO appears to be able to fulfill all the major goals set by the Open Research Challenge of the year 2022. 

    \subsection{Future Directions}
    
        Possible improvements for MARIO include:
        
        \begin{itemize}
            \item increasing the data-set created for the 2D pose estimation and the extraction of the 3D poses from the 2D poses;
            \item detecting additional actions such as walking, passing, and kicking;
            \item computing in real-time the recognition of the illegal defender foul;
            \item adding more statistics;
            \item improving the performance on low-end hardware. 
        \end{itemize}

%% file: code.tex
\section*{Addendum: Environment Setup and Codebase} \label{code}

    \textbf{Prerequisities.} MARIO has been tested on Ubuntu 20.04 LTS. \\
    
    \textbf{System Dependencies.} In order to resolve all the system dependendies, it is possibile to execute the following commands:
    
    \begin{verbatim}
        $ sudo apt install git python3-pip zlib1g-dev libjpeg-dev \ 
        libpng-dev
    \end{verbatim}
    
    \textbf{OpenCV.} In order to install OpenCV, you can execute the following command:
    
    \begin{verbatim}
        $ pip install opencv-python==4.6.0.66
    \end{verbatim}
    
    To install OpenCV from source, it is necessary to follow these steps:
    
    \begin{itemize}
        \item install the build tools and the dependencies:
        \begin{verbatim}
            $ sudo apt install build-essential cmake \ 
            pkg-config libgtk-3-dev libavcodec-dev \ 
            libavformat-dev libswscale-dev libv4l-dev \ 
            libxvidcore-dev libx264-dev libjpeg-dev libpng-dev \ 
            libtiff-dev gfortran openexr libatlas-base-dev \ 
            python3-dev python3-numpy libtbb2 libtbb-dev \ 
            libdc1394-22-dev libopenexr-dev \
            libgstreamer-plugins-base1.0-dev libgstreamer1.0-dev
        \end{verbatim}
        \item clone the OpenCV’s and OpenCV contrib repositories:
        \begin{verbatim}
            $ mkdir $HOME/opencv_build && cd $HOME/opencv_build

            $ git clone https://github.com/opencv/opencv.git

            $ git clone https://github.com/opencv/opencv_contrib.git
        \end{verbatim}
        \item create a temporary build directory, and navigate into it:
        \begin{verbatim}
            $ cd $HOME/opencv_build/opencv

            $ mkdir -p build && cd build
        \end{verbatim}
        \item setup the OpenCV build:
        \begin{verbatim}
            $ cmake -D CMAKE_BUILD_TYPE=RELEASE \
              -D CMAKE_INSTALL_PREFIX=/usr/local \
              -D INSTALL_C_EXAMPLES=ON \
              -D INSTALL_PYTHON_EXAMPLES=ON \
              -D OPENCV_GENERATE_PKGCONFIG=ON \
              -D OPENCV_EXTRA_MODULES_PATH=../opencv_contrib/modules \
              -D BUILD_EXAMPLES=ON ..
        \end{verbatim}
        \item start the compilation process:
        \begin{verbatim}
            $ make -j<number_of_cores>
        \end{verbatim}
        \item install OpenCV:
        \begin{verbatim}
            $ sudo make install
        \end{verbatim}
        \item verify the installation:
        \begin{verbatim}
            $  python3 -c "import cv2; print(cv2.__version__)"
        \end{verbatim}
    \end{itemize}

    \textbf{Python libraries.} Python modules dependencies are the following:
    
    \begin{verbatim}
        $ requests==2.28.0

        $ torch==1.11.0 to11.0 torchvision==0.12.0

        $ pyyaml==6.0

        $ tqdm==4.64.0

        $ matplotlib==3.5.2

        $ seaborn==0.11.2

        $ gdown==4.4.0

        $ cython==0.29.30

        $ tensorboard==2.9

        $ easydict==1.9

        $ scikit-learn==1.1.1

        $ protobuf==3.20.0

        $ https://github.com/KaiyangZhou/deep-person-reid/archive/master.zip
    \end{verbatim}
    
    \textbf{Clone and run the project.} In order to run the project, clone the repository with the following command:
    
    \begin{verbatim}
        $ sudo git clone https://github.com/unibas-wolves/MARIO.git
    \end{verbatim}
    
    and insert:
    
    \begin{itemize}
        \item the content of \href{https://drive.google.com/drive/folders/1hDBn8gZZ1LzGC5JKYuN2AIpA_oewmNM2?usp=sharing}{this} folder in \texttt{MARIO/detectionT};
        \item the content of \href{https://drive.google.com/drive/folders/1jHWJbsgEpoFRs8ttHWARSFuaemOJ4BsJ?usp=sharing}{this} folder in \texttt{MARIO/data};
        \item the content of \href{https://drive.google.com/drive/folders/1Uea9DB4tz7uAb36V6ydfzTGwpQrn3YzJ?usp=sharing}{this} folder in \texttt{MARIO/video};
    \end{itemize}
    
    To start the project, run the following command:
        
    \begin{verbatim}
        $ python3 ./gui/mario.py
    \end{verbatim}
    
    \textbf{Know problems.} Possible problems that may be encountered during a run of the application:
    
    \begin{itemize}
        \item \texttt{FileNotFoundError: No such file or directory: 'python'}
            \begin{itemize}
                \item \textbf{Solution}: change the value \texttt{python} in \texttt{python3} in \texttt{Preparation.py}, at line 70.
            \end{itemize}
        \item \texttt{FileExistsError: File exists: './MARIO/imbs-mt/images'}
            \begin{itemize}
                \item \textbf{Solution}: delete \texttt{images} folder in \texttt{MARIO/imbs-mt} folder.
            \end{itemize}
    \end{itemize}
    
    \subsection{Docker file}
        
        \textbf{Prerequisites.} Prerequisites includes Ubuntu 20.04 and Docker. \\
        
        \textbf{System Dependencies.} In order to resolve all the system dependencies, it is possible to execute the following commands:
    
        \begin{verbatim}
            $ sudo apt-get install ca-certificates curl gnupg \ 
            lsb-release
        \end{verbatim}
        
        \textbf{Import Docker’s official GPG key.} Import Docker GPG key using the following commands:
        
        \begin{verbatim}
            $ sudo mkdir -p /etc/apt/keyrings
            
            $ curl -fsSL https://download.docker.com/linux/ubuntu/gpg \
            | sudo gpg --dearmor -o /etc/apt/keyrings/docker.gpg
        \end{verbatim}
        
        \textbf{Setup Docker repository. } In order to setup the Docker repository, run the following commands:
        
        \begin{verbatim}
            $ echo "deb [arch=$(dpkg --print-architecture) \ 
            signed-by=/etc/apt/keyrings/docker.gpg]  \ 
            https://download.docker.com/linux/ubuntu \ 
            $(lsb_release -cs) stable" | sudo tee \ 
            /etc/apt/sources.list.d/docker.list > /dev/null
        \end{verbatim}
        
        \textbf{Install Docker Engine.} To install Docker Engine, run the following commands:
        
        \begin{verbatim}
            $ sudo apt-get install docker-ce docker-ce-cli containerd.io \ 
            docker-compose-plugin
        \end{verbatim}
        
        \textbf{Folder setup.} Clone the following repositories
        
        \begin{verbatim}
            $ sudo git clone --branch MARIO-docker \ 
            https://github.com/unibas-wolves/MARIO.git
            
            $ sudo git clone https://github.com/unibas-wolves/MARIO.git
            
            $ mv ./MARIO/ ./MARIO-docker/
        \end{verbatim}
        
        and insert:
    
        \begin{itemize}
            \item the content of \href{https://drive.google.com/drive/folders/1hDBn8gZZ1LzGC5JKYuN2AIpA_oewmNM2?usp=sharing}{this} folder in \texttt{MARIO-docker/MARIO/detectionT};
            \item the content of \href{https://drive.google.com/drive/folders/1jHWJbsgEpoFRs8ttHWARSFuaemOJ4BsJ?usp=sharing}{this} folder in \texttt{MARIO-docker/MARIO/data};
            \item the content of \href{https://drive.google.com/drive/folders/1Uea9DB4tz7uAb36V6ydfzTGwpQrn3YzJ?usp=sharing}{this} folder in \texttt{MARIO-docker/MARIO/video};
        \end{itemize}
        
        \textbf{Build and Setup Docker image.} In order to build and setup the Docker image, run the following commands:
        
        \begin{verbatim}
            $ sudo docker build -t robocup2022 .

            $ xhost +

            $ sudo docker run -it --volume=/tmp/.X11-unix:/tmp/.X11-unix \
            --device=/dev/dri:/dev/dri --env="DISPLAY=$DISPLAY" robocup2022

            $ cd src
        \end{verbatim}
        
        To start the project, run the following command:
        
        \begin{verbatim}
            $ python3 ./gui/mario.py
        \end{verbatim}
        
        \textbf{Known problems.} Possible problems that may be encountered when running the application include the following:
    
        \begin{itemize}
            \item \texttt{FileNotFoundError: No such file or directory: 'python'}
                \begin{itemize}
                    \item \textbf{Solution}: change the value \texttt{python} in \texttt{python3} in \texttt{Preparation.py}, at line 70.
                \end{itemize}
        \end{itemize}